\documentclass[10pt,twocolumn,letterpaper]{article}

\usepackage[pagenumbers]{cvpr} 

\definecolor{cvprblue}{rgb}{0.21,0.49,0.74}
\usepackage[pagebackref,breaklinks,colorlinks,allcolors=cvprblue]{hyperref}
\newcommand{\matdn}[2]{\mathbf{#1}_{\rm #2}}
\newcommand{\matup}[2]{\mathbf{#1}^{\rm #2}}
\newcommand{\matud}[3]{\mathbf{#1}^{\rm #2}_{\rm #3}}

\newcommand{\vecup}[2]{\boldsymbol{#1}^{\rm #2}}
\newcommand{\vecdn}[2]{\boldsymbol{#1}_{\rm #2}}
\newcommand{\vecupdown}[3]{\boldsymbol{#1}^{\rm #2}_{\rm #3}}
\newcommand{\myPara}[1]{\vspace{.05in}\noindent\textbf{#1}\quad}

\def\ourmodel{GFS-VL} 
\definecolor{red}{RGB}{234,0,0}
\definecolor{green}{RGB}{104,255,51}
\usepackage{multirow}
\usepackage{colortbl}
\usepackage{color, xcolor} 
\usepackage{subcaption}
\usepackage{listings}
\usepackage[ruled,linesnumbered]{algorithm2e}
\definecolor{mygray}{gray}{0.9}
\DeclareMathOperator*{\argmax}{arg\,max}

\def\paperID{} 
\def\confName{CVPR}
\def\confYear{2025}

\title{Generalized Few-shot 3D Point Cloud Segmentation \\ with Vision-Language Model}

\author{Zhaochong An$^{1,2}$, \quad Guolei Sun$^{2*}$, \quad Yun Liu$^3$\thanks{Corresponding authors: Guolei Sun and Yun Liu}, \quad Runjia Li$^4$, \quad Junlin Han$^4$,\\ 
Ender Konukoglu$^2$, \quad Serge Belongie$^1$\\
$^1$ Department of Computer Science, University of Copenhagen\\
$^2$ Computer Vision Laboratory, ETH Zurich\\
$^3$ College of Computer Science, Nankai University\\
$^4$ Department of Engineering Science, University of Oxford \\
}

\begin{document}
\maketitle
\begin{abstract}
Generalized few-shot 3D point cloud segmentation (GFS-PCS) adapts models to new classes with few support samples while retaining base class segmentation. 
Existing GFS-PCS methods enhance prototypes via interacting with support or query features but remain limited by sparse knowledge from few-shot samples. 
Meanwhile, 3D vision-language models (3D VLMs), generalizing across open-world novel classes, contain rich but noisy novel class knowledge. 
In this work, we introduce a \textbf{GFS}-PCS framework that synergizes dense but noisy pseudo-labels from 3D \textbf{VL}Ms with precise yet sparse few-shot samples to maximize the strengths of both, named~\textbf{\ourmodel}. 
Specifically, we present a prototype-guided pseudo-label selection to filter low-quality regions, followed by an adaptive infilling strategy that combines knowledge from pseudo-label contexts and few-shot samples to adaptively label the filtered, unlabeled areas. 
Additionally, we design a novel-base mix strategy to embed few-shot samples into training scenes, preserving essential context for improved novel class learning.
Moreover, recognizing the limited diversity in current GFS-PCS benchmarks, we introduce two challenging benchmarks with diverse novel classes for comprehensive generalization evaluation. 
Experiments validate the effectiveness of our framework across models and datasets. 
Our approach and benchmarks provide a solid foundation for advancing GFS-PCS in the real world. The code is at \href{https://github.com/ZhaochongAn/GFS-VL}{here}.

\end{abstract}

\begin{figure}[t!]
    \centering
    \includegraphics[width=.95\linewidth]{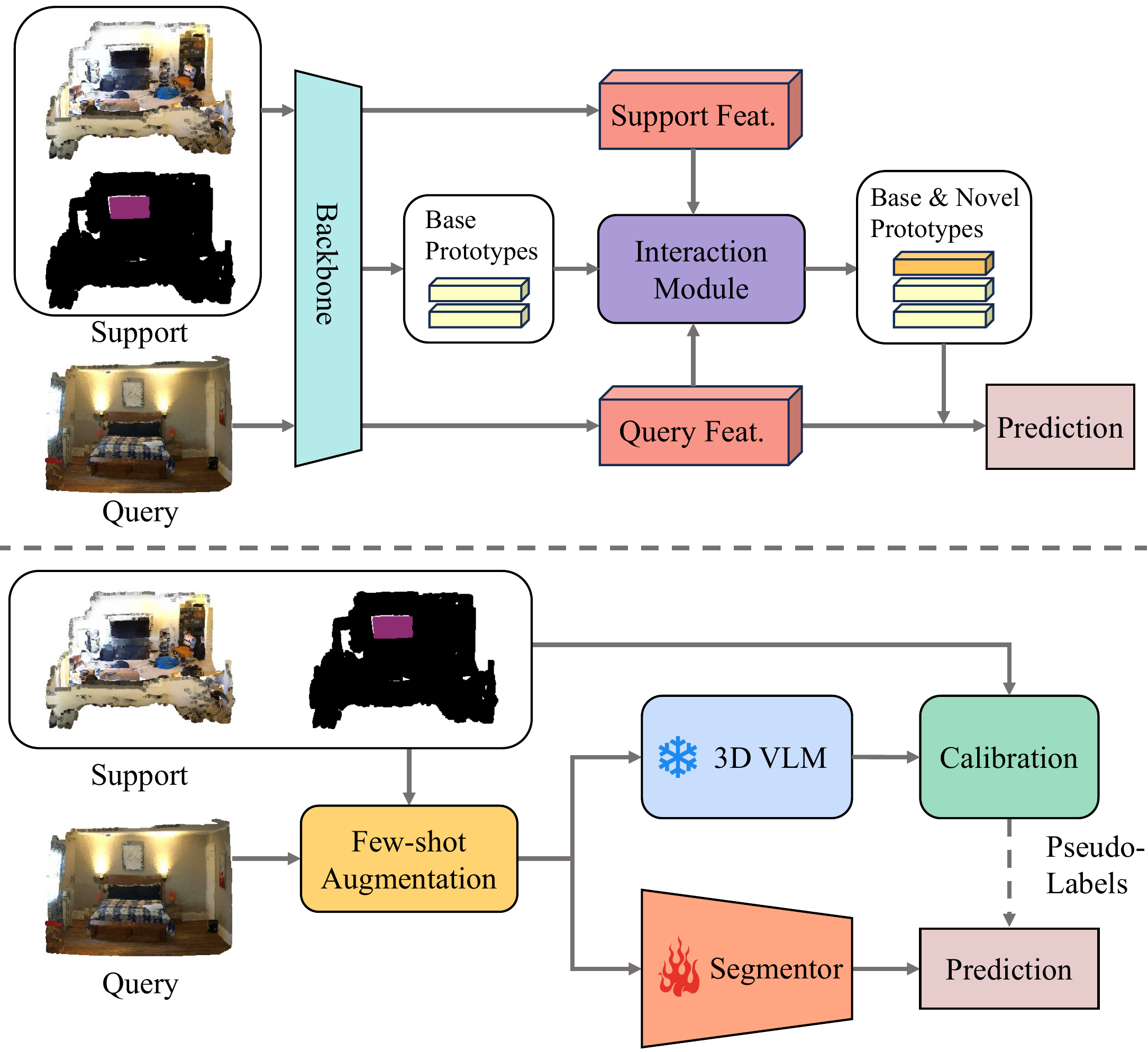}
    \vspace{-.10in}
    \caption{
    \textbf{Comparison of our framework with previous work.} 
    \textit{Top:} Prior work~\cite{tian2022generalized,xu2023generalized} primarily enhances prototypes through interaction modules that integrate support/query features, making predictions based on refined prototypes. However, they are limited by the sparse knowledge from few-shot samples.
    \textit{Bottom:} Our framework addresses this limitation by leveraging the extensive open-world knowledge from 3D VLMs through pseudo-labels. We mitigate the noise inherent in 3D VLMs by calibrating their raw pseudo-labels with precise few-shot samples, thereby effectively expanding novel class knowledge while ensuring reliability.
    }
    \label{fig:intro}
    \vspace{-.15in}
\end{figure}

\section{Introduction}
\label{sec:intro}
Understanding dense 3D semantics is essential for many vision applications~\cite{liu2024lion,jiang2024pcotta,sun2024point,pan2024distribution,zhang2024voxel,sun2024review,wang2024panoptic}, and \textit{few-shot point cloud semantic segmentation} (FS-PCS) has emerged as a valuable task~\cite{zhao2021few,an2024rethinking}, enabling models to extend from base to novel classes with minimal annotations for novel classes.
However, typical few-shot models require additional support samples for each novel class at inference and only predict novel classes, ignoring base classes. 
To address this, \textit{generalized few-shot point cloud semantic segmentation} (GFS-PCS)~\cite{xu2023generalized} was introduced,
allowing models to directly segment both base and novel classes after few-shot adaptation, making it more practical for real-world use.

Current GFS-PCS models~\cite{liu2023learning, hossain2024visual, liu2024harmonizing,xu2023generalized} primarily utilize prototype learning~\cite{snell2017prototypical}, representing each class as a prototype and predicting based on the relationship between query points and these prototypes. 
As shown in~\cref{fig:intro},
these methods mainly focus on refining prototypes through interactions with support/query samples for enhanced segmentation. 
For instance, CAPL~\cite{tian2022generalized} adapts base prototypes to novel classes using co-occurrence knowledge from support samples and contextual information from queries.  GW~\cite{xu2023generalized} encodes shared geometric structures into geometry prototypes to enhance semantic prototypes.  
However, these approaches remain limited in novel class generalization due to the sparse knowledge available from few-shot samples.

In parallel, \textit{3D vision-language models} (3D VLMs) have been developed to enable open-vocabulary recognition by aligning 3D and language features. 
Leveraging language models trained on vast open-text data, 
3D VLMs exhibit strong generalization abilities, allowing recognition of open-set classes in 3D. 
Since paired 3D-text data is scarce, some approaches~\cite{chen2023clip2scene, huang2023clip2point, peng2023openscene, takmaz2023openmask3d, zhang2023clip} distill 2D features from 2D VLMs~\cite{ghiasi2022scaling, li2022language} into their 3D encoders, while others~\cite{ding2023pla, jiang2024open, ding2024lowis3d, yang2024regionplc} use captioning models~\cite{rennie2017self, wang2022ofa} to generate scene- or region-level descriptions, enabling point-language alignment for direct 3D learning. 
Recognizing the open-world potential of 3D VLMs, we propose utilizing their rich knowledge to enhance GFS-PCS.
A straightforward method to integrate 3D VLMs~\cite{peng2023openscene,yang2024regionplc} is to generate dense pseudo-labels of novel classes as additional supervision. 
However, predictions from 3D VLMs are often noisy, compounding errors in GFS-PCS models. 
Meanwhile, sparse support samples offer accurate annotations for novel classes.
Therefore, given the dense but noisy pseudo-labels from 3D \textbf{VL}Ms and the accurate yet limited support samples, we propose a new \textbf{GFS}-PCS framework, named~\textbf{\ourmodel}, to combine the strengths of both, as in~\cref{fig:intro}.

Specifically, \ourmodel~incorporates three novel techniques.
First, 
we introduce a pseudo-label selection technique that uses accurate few-shot data to filter pseudo-labels, retaining only high-quality regions while excluding noisy predictions. 
Second, 
as filtered wrong predictions will leave some regions unlabeled, potentially corresponding to novel objects, we present an adaptive infilling approach to enrich these regions. It combines knowledge from pseudo-label contexts and few-shot samples to construct an adaptive prototype set to label unlabeled regions, effectively considering both the completion of incomplete masks and the discovery of missing classes.
Third, 
to further utilize few-shot samples, we propose a novel-base mix strategy, embedding support samples into training scenes. 
Unlike traditional 3D data augmentation~\cite{nekrasov2021mix3d,xiao2022polarmix,saltori2023compositional,zhu2023curricular,zhan2023real,wu2024mitigating}, which mainly aims at fully-supervised segmentation and mixes object contexts, 
our approach emphasizes preserving contextual cues, which is crucial for novel class learning~\cite{tian2022generalized} by helping identify challenging novel classes.

Furthermore, we identify limitations in current evaluation benchmarks. 
Existing benchmarks based on ScanNet~\cite{dai2017scannet} and S3DIS~\cite{armeni20163d} datasets include only six novel classes, limiting diversity and failing to represent the complexity of real-world scenarios where novel classes are constantly varying. 
To address this, we introduce two challenging benchmarks: one with 40 novel classes from ScanNet200~\cite{rozenberszki2022language} and another with 18 novel classes from ScanNet++~\cite{yeshwanth2023scannet++}. 
As detailed in~\cref{sec:newsetting}, these benchmarks provide broader and more representative coverage of novel classes, enabling a more comprehensive evaluation of models' generalization capabilities.

By fully integrating the benefits of 3D VLMs and few-shot data, our approach achieves state-of-the-art GFS-PCS performance on both existing and newly established benchmarks.
Experiments demonstrate the effectiveness and generalizability of our framework across various models and datasets. Additionally, previous baselines exhibit limited performance when evaluated on our new benchmarks, underscoring the necessity of our benchmarks for assessing real-world generalization. Together, our methods and benchmarks would offer critical insights and tools to advance future research on GFS-PCS.

\section{Related Work}
\subsection{Few-shot 3D Point Cloud Segmentation}
3D point cloud segmentation is fundamental in understanding scene semantics, with many fully-supervised methods advancing this field~\cite{park2022fast, nie2022pyramid, lai2022stratified,li2023hierarchical, zhang2023improving, wu2024point, kolodiazhnyi2024oneformer3d, wang2024gpsformer, han2024subspace, zheng2023spherical,deng2024linnet}. 
However, these methods need expensive large-scale point-level annotations and have fixed output spaces.
To address these limitations, FS-PCS was introduced in attMPTI~\cite{zhao2021few}, aiming to generalize to novel classes with limited support samples. 
FS-PCS research can be divided into two categories based on how relationships between query points and support classes are modeled:
i) Feature optimization~\cite{he2023prototype, ning2023boosting, zhu2023cross, li2024localization, mao2022bidirectional, wang2023few, zhang2023few, zhu2024no,wei2024gandpfewshot} -- 
These methods refine support prototypes or query features to enhance class separation. Final predictions are made using non-parametric, distance-based metrics, which implicitly model support-query relationships. 
ii) Correlation optimization~\cite{an2024rethinking,an2024multimodality} -- 
These approaches directly optimize correlations between support and query samples, explicitly modeling their relationships. 
COSeg \cite{an2024rethinking} pioneered this approach recently and corrected two issues of foreground leakage and sparse point distribution in the previous FS-PCS setting.

\subsection{Generalized Few-shot 3D Point Cloud Segmentation}
While standard few-shot models adapt effectively to novel classes with limited data, they are constrained to predict only novel classes and require support samples to specify target classes during inference. 
A more practical task, generalized few-shot segmentation, occurs to require predicting both base and novel classes at inference without support samples, as first introduced in 2D segmentation~\cite{cermelli2021prototype,tian2022generalized,myers2021generalized,lang2022learning,lu2023prediction,hajimiri2023strong,liu2023learning,hossain2024visual,liu2024harmonizing}.
The pioneer work PIFS~\cite{cermelli2021prototype}, using the prototype learning paradigm~\cite{snell2017prototypical}, fine-tunes base and novel prototypes with a distillation loss, while CAPL~\cite{tian2022generalized} refines base prototypes using co-occurrence priors from support samples and dynamic contextual information from queries. 
For 3D, GW~\cite{xu2023generalized} introduces this setting to point cloud segmentation. 
GW models shared local geometric structures across base and novel classes as ``geometry words'' and then builds geometric prototypes to enhance the semantic prototypes, which are learned similarly to CAPL~\cite{tian2022generalized} by leveraging contextual information.
To mine background semantics, Tsai \etal \cite{tsaipseudo} clustered background points to generate pseudo-class prototypes distinct from base classes, leveraging multiple 2D views and 2D foundation models to link these points with class prompts.

\subsection{3D Vision-Language Models}
3D VLMs align 3D point cloud features with language features, enabling open-world 3D understanding. 
However, developing these models poses unique challenges compared to 2D, mainly due to the scarcity of paired 3D-text data. To address this, recent work leverages multi-view 2D images, commonly associated with 3D point clouds, as intermediaries~\cite{li2024dense}. Some methods~\cite{chen2023clip2scene,huang2023clip2point,jatavallabhula2023conceptfusion,peng2023openscene,takmaz2023openmask3d,wang2022ofa,zhang2023clip,zhou2024point} distill 2D features from 2D VLMs~\cite{ghiasi2022scaling,li2022language,radford2021learning,yao2022detclip} into their 3D encoders. 
For instance, OpenScene~\cite{peng2023openscene} aligns 3D and text representations by optimizing 3D-2D alignment using 2D VLMs~\cite{ghiasi2022scaling,li2022language}. 
However, extracting these 2D features is computationally expensive, and the learned 3D features may inherit 2D prediction errors.
Other approaches~\cite{ding2023pla,jiang2024open,ding2024lowis3d} generate point-language paired data by using captioning models~\cite{rennie2017self,wang2022ofa} to produce text descriptions of images. 
While effective, these captions are often at the scene level, limiting their ability to capture fine-grained 3D features. 
RegionPLC~\cite{yang2024regionplc} recently introduced high-quality region-level 3D-language associations, supporting robust 3D learning by dense regional language supervision.

\subsection{Point Cloud Data Augmentation}
To address data limitations in the 3D domain, numerous methods have been developed to expand point cloud distributions. 
One category augments individual point clouds by altering geometric properties~\cite{li2020pointaugment, sheshappanavar2021patchaugment, kim2021point, liu2023hierarchical, qiu2024leveraging}, using techniques such as shape transformations and patch shuffling. 
Another approach uses mixing techniques for 3D objects~\cite{chen2020pointmixup, lee2021regularization, zhang2022pointcutmix} and 3D scenes~\cite{fang2021lidar, xiao2022polarmix, saltori2023compositional, zhu2023curricular, zhan2023real, wu2024mitigating}. For instance, Mix3D~\cite{nekrasov2021mix3d} mixes points from two scenes as an out-of-context augmentation for semantic segmentation, while methods like~\cite{saltori2023compositional} select domain-specific points with semantic information to mix across domains. 
Other methods, such as~\cite{wu2024mitigating}, relocate objects to less frequent locations to vary spatial distributions.
Notably, most augmentation methods in semantic segmentation~\cite{nekrasov2021mix3d, wu2024mitigating} and detection~\cite{zhu2023curricular} modify the original context of target classes, pushing models to learn object patterns independently of surroundings. 
However, we argue that preserving contextual dependencies is crucial for GFS-PCS, where novel classes often involve challenging, hard-to-detect objects. 
Isolated object patterns alone are insufficient for effective novel class generalization~\cite{tian2022generalized}. In contrast, our proposed novel-base mix augmentation retains key contextual information when incorporating support samples into training scenes, enhancing models' ability to recognize novel classes.

\section{GFS-PCS Overview}
\begin{figure*}[!ht]
    \centering
    \includegraphics[width=.95\linewidth]{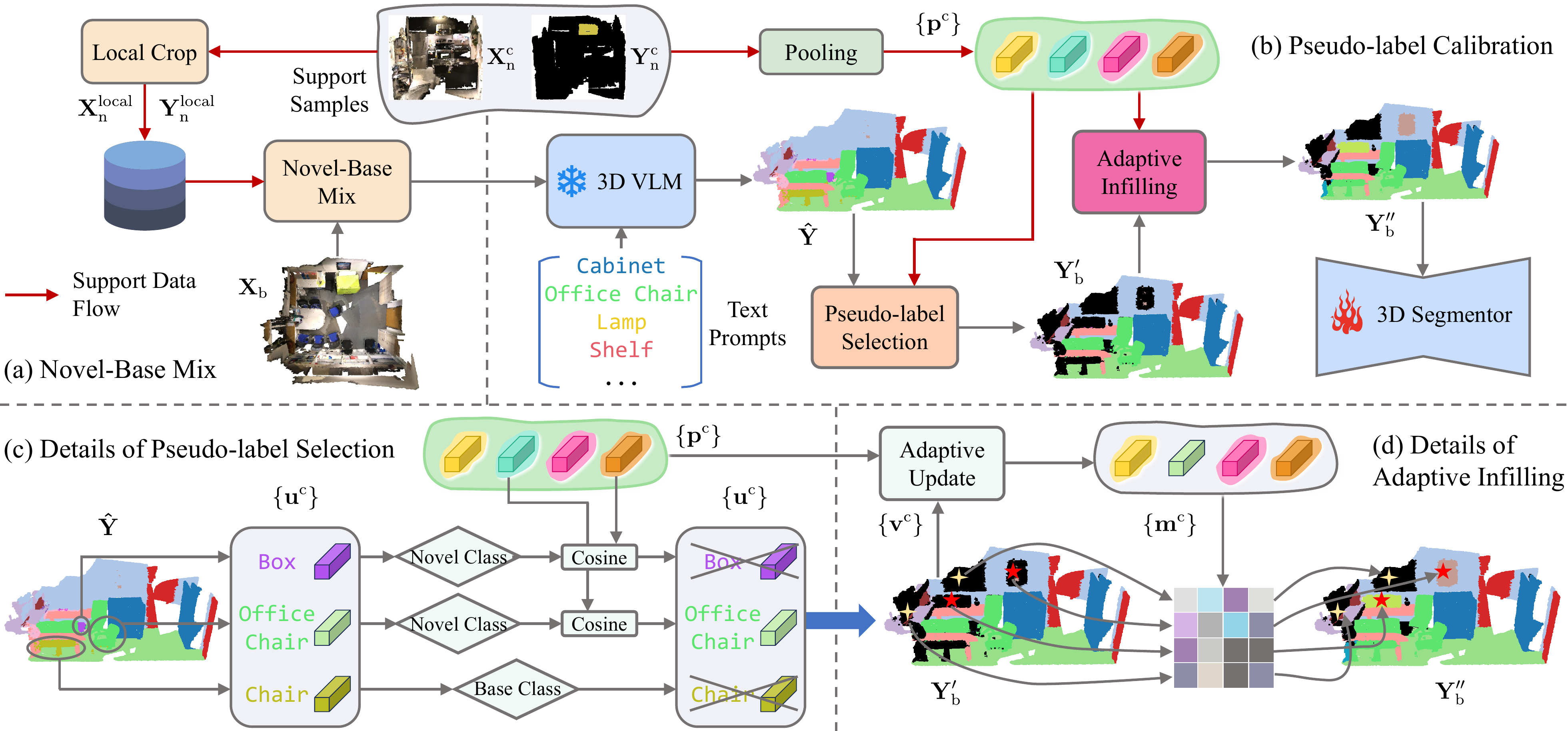}
    \vspace{-0.05in}
    \caption{
    \textbf{Overview of the proposed~\ourmodel.}
    (a), (b) Given an input point cloud $\matdn{X}{b}$, we apply a novel-base mix to embed support samples into the training scene while preserving essential context. 
    The scene is then processed by a 3D VLM, using all class names as prompts to generate raw predictions $\mathbf{\hat{Y}}$. 
    Leveraging support prototypes $\{\matup{p}{c}\}$, the raw predictions undergo pseudo-label selection to filter out noisy regions, followed by adaptive infilling to label the filtered, unlabeled areas, yielding refined supervision $\matud{Y}{\prime\prime}{b}$ for training the 3D segmentor. (c), (d) illustrate the details of the pseudo-label selection and adaptive infilling processes.
    }
    \label{fig:arch}
    \vspace{-0.20in}
\end{figure*}

\subsection{Problem Definition}
\label{sec:formatting}

GFS-PCS requires the model to identify both base and novel classes present in test scenes.
Let $\vecup{C}{b}$ denote the set of base class names with size $N_b$ and $\vecup{C}{n}$ the set of novel class names with size $N_n$. The total number of classes is $N_c = N_b + N_n$. The base and novel class sets are mutually exclusive, \ie, $\vecup{C}{b} \cap \vecup{C}{n} = \varnothing$. Evaluation is conducted on the test dataset $\matdn{D}{test}$ to assess segmentation across all classes $\vecup{C}{b} \cup \vecup{C}{n}$.
For training, the model is first trained on base classes and then registered with novel classes.
The base dataset $\matdn{D}{base}$ is defined as $\matdn{D}{base} = \{\matud{X}{i}{b}, \matud{Y}{i}{b}\}_{i}$, where $\matud{X}{i}{b}$ represents the $i$-th point cloud and $\matud{Y}{i}{b}$ contains the corresponding base class labels. 
For the novel class dataset, we define $\matdn{D}{novel}$ as the collection of $K$-shot support samples: $\big\{\{\matud{X}{c}{k}, \matud{Y}{c}{k}\}_{k=1}^K\big\}_{c=N_b}^{N_c}$. Each novel class $c$ has $K$ support samples $\{\matud{X}{c}{k}\}_{k=1}^K$ with exclusive labels $\{\matud{Y}{c}{k}\}_{k=1}^K$. 
The base classes occupy class indices in the range $[0, N_b)$, while novel classes are indexed in the range $[N_b, N_c)$, with background regions labeled as $-1$.
For simplicity, we denote a data sample from $\matdn{D}{base}$ as $\matdn{X}{b}$ and $\matdn{Y}{b}$, while $\matud{X}{c}{n}$ and $\matud{Y}{c}{n}$ refer to a support sample of class $c$ in $\matdn{D}{novel}$.

\begin{table}[t]
\centering
\renewcommand{\arraystretch}{.95}
\resizebox{\linewidth}{!}{
\begin{tabular}{lcccccc}
\toprule
Dataset & Base & Novel  & Max (F) & Min (F)    & Max (P) & Min (P)   \\  \midrule
S3DIS & 7 & 6  & 185 & 29    & 59,929 & 30,013    \\ 
ScanNet & 13 & 6  & 411 & 133   & 4,479 & 1,148   \\ 
\rowcolor{orange!20} ScanNet200 & 12 & 45  & \textbf{733} & \textbf{102}    & 12,641 & 279   \\  
\rowcolor{orange!20} ScanNet++ & 12 & 18  & 143 & 82    & \textbf{84,375} & \textbf{604}   \\  
\bottomrule
\end{tabular}
}
\vspace{-2mm}
\caption{Statistics for GFS-PCS benchmarks across four datasets. Base/Novel indicates the number of base and novel classes. Max/Min (F) is the maximum and minimum occurrences of each novel class across the entire dataset, while Max/Min (P) is the maximum and minimum average number of points per novel class.}
\label{dataset}
\vspace{-.1in}
\end{table}

\subsection{New Evaluation Benchmarks}
\label{sec:newsetting}
Current evaluation benchmarks for GFS-PCS utilize the ScanNet~\cite{dai2017scannet} and S3DIS~\cite{armeni20163d} datasets. However, we identify limitations in the number and diversity of novel classes in these benchmarks. 
As shown in~\cref{dataset}, while ScanNet includes 13 base classes and S3DIS has 7, both datasets include only 6 novel classes.
Our analysis of the frequency and the average number of points per novel class further highlights the lack of diversity among these limited novel classes in these two benchmarks.
This indicates that current evaluations are not sufficiently representative of the complexity of real-world novel categories, failing to robustly assess models' generalization abilities.
To address this, we propose two new and more challenging evaluation benchmarks based on ScanNet200~\cite{rozenberszki2022language} and ScanNet++~\cite{yeshwanth2023scannet++} datasets. These new benchmarks, as detailed in~\cref{dataset}, feature a larger number and greater diversity of novel classes, providing a more realistic and comprehensive testbed for evaluating model generalization to novel categories.

\section{Method}
\label{sec:method}
\subsection{Overview}
In our proposed framework~\ourmodel, we adopt a canonical segmentor architecture consisting of a backbone and a linear classification head. 
This minimalistic structure retains flexibility and simplicity, facilitating reproducibility.

Following the standard GFS-PCS procedure~\cite{xu2023generalized}, the segmentor is initially trained on base classes. The backbone, denoted as $\Phi$, and the linear classifier for base classes, $\mathcal{H}_b$, are employed to make base class predictions $\matdn{P}{b}$: 
\begin{align}
\label{eq:cs}
\mathbf{F} = \Phi(\matdn{X}{b}) \in \mathbb{R}^{N_p \times D}, \matdn{P}{b} = \mathcal{H}_b(\mathbf{F}) \in \mathbb{R}^{N_p \times N_b}, 
\end{align}
where $N_p$ is the number of points in the point cloud input $\matdn{X}{b}$, and $\mathbf{F}$ represents backbone features with a channel dimension of $D$.
Then, when registering novel classes, a new linear classifier $\mathcal{H}_n$ is introduced to handle the novel classes. 
The concatenated predictions from both classifiers form the complete output $\mathbf{P}$ for all base and novel classes:
\begin{align}
\mathbf{P} = [\mathcal{H}_b(\mathbf{F}), \mathcal{H}_n(\mathbf{F})] \in \mathbb{R}^{N_p\times N_c}, 
\end{align}
where $[\cdot,\cdot]$ denotes concatenation.
Subsequently, the model is fine-tuned on the base and few-shot samples, enabling it to simultaneously segment both base and novel classes.
To fully exploit the rich knowledge from 3D VLMs while minimizing potential noise interference, we propose~\ourmodel, as shown in~\cref{fig:arch}, to maximize the utility of limited but accurate novel samples to guide the learning process.
We detail each designed module in the following sections, presented under the 1-shot setting for clarity.

\subsection{Pseudo-label Selection}
A direct method for utilizing 3D VLMs in GFS-PCS is to use their predictions as pseudo-labels for novel classes. 
However, these raw predictions are noisy, hindering few-shot models from learning good novel class representations. 
Moreover, such noisy pseudo-labels could introduce error accumulation from the 3D VLMs into the few-shot models.

Therefore, we introduce an effective pseudo-label selection method by utilizing the valuable few-shot support samples to guide the selection of reliable novel class predictions.
Specifically, for each novel class, we first compute support prototypes using the few-shot samples. 
This is achieved by applying the vision encoder $\Theta_v$ of the 3D VLM $\Theta$ to compute masked average features for each novel class: 
\begin{align}
\label{eq:pc}
\begin{split}
\matud{F}{c}{n} = \Theta_v(\matud{X}{c}{n}) &\in \mathbb{R}^{N_p \times D_v}, \\
\matup{p}{c} = \frac{\sum_{i=1}^{N_p} \matud{F}{c}{n,i} \matud{Y}{c}{n,i}}{\sum_{i=1}^{N_p} \matud{Y}{c}{n,i}}, \qquad& c=N_b,\dots,N_c-1,\\
\end{split}
\end{align}
where $D_v$ is the feature dimension of the 3D VLM, and $\matup{p}{c} \in \mathbb{R}^{D_v}$ represents the support prototype for novel class $c$. 
For clarity, we define this prototype extraction process as $\mathcal{F}_{{\rm pool}}$. Thus,~\cref{eq:pc} becomes: $\matup{p}{c} = \mathcal{F}_{{\rm pool}}(\matud{X}{c}{n}, \matud{Y}{c}{n})$.

Next, given the current base training input $\matdn{X}{b}$ with base labels $\matdn{Y}{b}$, 
we prompt the 3D VLM $\Theta$ using all base and novel class names to obtain predictions $\mathbf{\hat{Y}} \in \mathbb{R}^{N_p}$. 
Let $\vecdn{\hat{C}}{n}$ be the novel class indices existing in $\mathbf{\hat{Y}}$.
We then compute the predicted prototype $\matup{u}{c}$ for each novel class in $\vecdn{\hat{C}}{n}$: 
\begin{align}
\begin{split}
\matup{u}{c} = \mathcal{F}_{{\rm pool}}(\matdn{X}{b}, \mathbf{1}_{[\mathbf{\hat{Y}} = c]}), &\qquad c \in \vecdn{\hat{C}}{n}.\\
\end{split}
\end{align}
Here, $\mathbf{1}_{[\mathbf{\hat{Y}} = c]}$ is a binary mask, set to 1 where $\mathbf{\hat{Y}}$ equals the class index $c$ and to 0 otherwise.
Then, we can filter the raw predictions to select high-quality novel class pseudo-labels:
\begin{align}
\matdn{\hat{Y}}{i} = \begin{cases} 
-1, & \text{if } \matdn{\hat{Y}}{i} \in [0, N_b) \text{ or } \\
    &\hspace{1.5mm} \bigl(\matdn{\hat{Y}}{i} \in [N_b, N_c) \text{ and } \\
&\hspace{0.8mm} \quad \text{sim}(\matup{u}{\matdn{\hat{Y}}{i}}, \matup{p}{\matdn{\hat{Y}}{i}}) < \tau\bigr), \\
\matdn{\hat{Y}}{i}, & \text{otherwise}.
\end{cases}
\end{align}
Here, if the predicted label $\matdn{\hat{Y}}{i}$ for the $i$-th point is a base class, or a novel class with cosine similarity below a threshold $\tau$ between the predicted class prototype $\matup{u}{\matdn{\hat{Y}}{i}}$ and the support prototype $\matup{p}{\matdn{\hat{Y}}{i}}$, we filter this pseudo-label by setting it to $-1$. Otherwise, we retain the original pseudo-label.
Note this filtering process can be efficiently implemented using mask-based indexing without iterating each point.

Now the updated $\mathbf{\hat{Y}}$ contains only reliable pseudo-labels for novel classes.
Given the original base class labels $\matdn{Y}{b}$, its background region (labeled as $-1$) serves as a potential area for novel classes.
Therefore, we merge the updated $\mathbf{\hat{Y}}$ into the background region in $\matdn{Y}{b}$ to generate augmented labels $\matud{Y}{\prime}{b}$ with additional reliable novel class pseudo-labels: 
\begin{align}
\begin{split}
\matud{Y}{\prime}{b} &= \matdn{Y}{b}, \\
\matud{Y}{\prime}{b}[\matud{Y}{\prime}{b} = -1] &= \mathbf{\hat{Y}}[\matud{Y}{\prime}{b} = -1].
\end{split}
\end{align}

\subsection{Adaptive Infilling}
After selection, $\matud{Y}{\prime}{b}$ includes reliable supervision for novel classes, while some regions remain unlabeled due to filtered low-quality predictions.
These filtered predictions from the 3D VLM usually assign wrong labels, 
either by entirely missing true novel areas or partially mislabeling them~\cite{thyagharajan2022segment}. 
Consequently, $\matud{Y}{\prime}{b}$ contains unlabeled regions that potentially correspond to missing or incomplete novel labels.

To address these gaps, we propose an adaptive infilling approach that utilizes both the few-shot samples and the current labels $\matud{Y}{\prime}{b}$ to build an adaptive prototype set for novel classes.
This set allows us to assign novel labels adaptively to unlabeled regions, ensuring more comprehensive coverage of novel classes.
We begin by extracting novel class prototypes from $\matud{Y}{\prime}{b}$: 
\begin{align}
\matup{v}{c} = \mathcal{F}_{{\rm pool}}(\matdn{X}{b}, \mathbf{1}_{[\matud{Y}{\prime}{b} = c]}), \quad c\in \vecupdown{C}{y}{n},
\end{align}
where $\vecupdown{C}{y}{n}$ represents the novel class indices present in $\matud{Y}{\prime}{b}$. 
Using both these extracted prototypes and pre-computed support prototypes from the few-shot samples, we construct an adaptive prototype set, defined as $\{\matup{m}{c}\}$, where:
\begin{align}
\matup{m}{c} = \begin{cases}
    \matup{v}{c}, & \text{if } c \in \vecupdown{C}{y}{n}, \\
    \matup{p}{c}, & \text{otherwise},
\end{cases}
\quad \text{for } c = N_b, \dots, N_c-1.
\end{align}
This set $\{\matup{m}{c}\}$ incorporates novel class prototypes $\matup{v}{c}$ from $\matud{Y}{\prime}{b}$ if they exist; otherwise, it defaults to the few-shot support prototypes $\matup{p}{c}$.
By adapting to the current pseudo-labels, this set facilitates the completion of incomplete novel class pseudo-labels while allowing for the discovery of missed novel classes based on support prototypes.

Next, we initialize $\matud{Y}{\prime\prime}{b} = \matud{Y}{\prime}{b}$. For each unlabeled point $\matud{Y}{\prime\prime}{b,i}$ with feature $\matdn{F}{b,i}$ from $\Theta_v$, we calculate its cosine similarity with each prototype $\matup{m}{c}$ as $S_{\rm b,i}^{\rm c} = \text{sim}(\matdn{F}{b,i}, \matup{m}{c})$
and assign the corresponding novel class label if the maximum similarity exceeds a threshold $\delta$:
\begin{align}
\matud{Y}{\prime\prime}{b,i} =  \begin{cases}
\argmax\limits_{c}\, S_{\rm b,i}^{\rm c}, & \text{if } \max\limits_{c}\, S_{\rm b,i}^{\rm c} \geq \delta, \\
-1, & \text{otherwise}.
\end{cases}
\end{align}
This adaptive infilling mechanism effectively integrates knowledge from few-shot support samples with the current pseudo-label context, creating adaptive prototypes that help discover missed novel objects and complete partial pseudo-labels, 
thereby enhancing the quality of novel region labels.

\subsection{Novel-Base Mix}
To more sufficiently utilize support samples, we introduce a novel-base mix approach that effectively integrates these valuable samples with the training data. 
Specifically, we start by randomly sampling a novel sample $\matud{X}{c}{n}$ from $\matdn{D}{novel}$.
To enhance the model's focus on novel class, we crop the local bounding region based on the novel class mask $\matud{Y}{c}{n}$:
\begin{align}
\begin{split}
\matud{X}{local}{n}, \matud{Y}{local}{n} = \mathcal{F}_{{\rm crop}}(\matud{X}{c}{n}, \matud{Y}{c}{n}),  \\
\end{split}
\end{align}
where $\mathcal{F}_{{\rm crop}}$ represents the local cropping operation.

We then construct a new training input by mixing the cropped novel sample $\matud{X}{local}{n}$ with the current base input $\matdn{X}{b}$. 
Unlike previous mixup methods~\cite{nekrasov2021mix3d,wu2024mitigating}, which discard scene context, 
we argue that context information is essential for models to better recognize challenging novel objects and propose to preserve it.
To achieve this, we extract the four corners in the XY plane for both $\matud{X}{local}{n}$ and $\matdn{X}{b}$, and then select a pair of opposite corners between them: 
\begin{align}
\begin{split}
\matdn{L}{b}, \matud{L}{local}{n} = \mathcal{F}_{{\rm pair}}(\mathcal{F}_{{\rm corner}}(\matdn{X}{b}), \mathcal{F}_{{\rm corner}}(\matud{X}{local}{n})), \\
\end{split}
\end{align}
where $\mathcal{F}_{{\rm corner}}$ extracts the four corner points in the XY projections, and $\mathcal{F}_{{\rm pair}}$ randomly selects an opposing corner pair. 
Possible pairs include, for example, the leftmost corner of $\matdn{X}{b}$ with the rightmost corner of $\matud{X}{local}{n}$, or the uppermost corner of $\matdn{X}{b}$ with the lowermost corner of $\matud{X}{local}{n}$. 
Here, $\matdn{L}{b}$ and $\matud{L}{local}{n}$ are the coordinates of the selected corners.
To align $\matdn{X}{b}$ with $\matud{X}{local}{n}$ at the chosen corners, we compute a translation vector $\mathbf{T} = \matdn{L}{b} - \matud{L}{local}{n}$ and apply it to translate $\matud{X}{local}{n}$, yielding the final mixed result. 
This ensures a close connection between the samples without losing context, which is crucial for effective novel class learning. 
Visualizations of the output can be found in~\cref{sec:abl}, with further details in the supplementary material.

\section{Experiments}

\subsection{Experimental Setup}
\label{sec:expsetup}
\noindent\textbf{Datasets.}
Our new evaluation benchmark builds on two datasets: 1) 
ScanNet200~\cite{rozenberszki2022language} -- This dataset extends the labeling scope of ScanNet~\cite{dai2017scannet} from 20 to 200 classes, adding finer-grained subclasses of existing categories and numerous new classes.
2) 
ScanNet++~\cite{yeshwanth2023scannet++} -- Comprising 460 scenes with annotations for over 1000 unique classes, ScanNet++ is designed to capture a wide range of object types.
To establish a comprehensive GFS-PCS benchmark, 
we selected the most frequent classes from each dataset, ensuring adequate representation across scenes. 
Our final benchmark includes 57 classes for ScanNet200 (with 40 novel classes) and 30 classes for ScanNet++ (with 18 novel classes). 
Full class lists are in the supplementary material.
We follow the standard training/testing splits for each dataset, adhering to the preprocessing and augmentation settings from~\cite{wu2024point}, where raw input points are voxelized at a 0.02m grid size. 
Notably, unlike prior GFS-PCS evaluations~\cite{xu2023generalized, zhao2021few}, which test models on small blocks, we test on whole scenes to better simulate real-world scenarios.

\noindent\textbf{Implementation Details.} 
Our framework uses a straightforward segmentor with a backbone and a linear classification head, optimized for efficiency and simplicity. 
By default, we use Point Transformer V3 (PTv3)~\cite{wu2024point} as the backbone and 3D VLM RegionPLC~\cite{yang2024regionplc}.
The segmentor is first pre-trained on base classes of each dataset, after which we add a separate linear classification head for novel classes, enabling lightweight and efficient adaptation to novel classes in 20 fine-tuning epochs.
The pre-training setting follows~\cite{wu2024point} with 800 epochs, and we use Adam optimizer for fine-tuning with learning rates: 0.001 for ScanNet200 and ScanNet, and 0.007 for ScanNet++. More details are in the supplementary material.
For evaluation, we adopt metrics outlined in~\cite{xu2023generalized}: mean Intersection-over-Union (mIoU) for base classes (mIoU-B), novel classes (mIoU-N), all classes (mIoU-A), and the harmonic mean of mIoU-B and mIoU-N (HM) which captures overall performance while mitigating bias towards base classes~\cite{ye2021learning}.

\begin{table*}[t]
\centering
\renewcommand{\arraystretch}{.95}
\newcommand{\gray}{\cellcolor{mygray}}
\resizebox{0.98\linewidth}{!}{
\begin{tabular}{p{2.5cm}<\raggedright |p{1.5cm}<\centering p{1.5cm}<\centering p{1.5cm}<\centering p{1.5cm}<\centering p{0.01cm}<\centering p{1.5cm}<\centering p{1.5cm}<\centering p{1.5cm}<\centering p{1.5cm}<\centering}
\toprule
\multirow{2}{*}{Method} & \multicolumn{4}{c}{5-shot}  & \phantom{a} & \multicolumn{4}{c}{1-shot}   \\ \cmidrule{2-5} \cmidrule{7-10} 
                        & mIoU-B & mIoU-N & mIoU-A & HM   & \phantom{a} & mIoU-B & mIoU-N & mIoU-A & HM \\ \midrule
Fully Supervised        & 68.70  & 39.32  & 45.51  & 50.02 & \phantom{a} & 68.70  & 39.32  & 45.51  & 50.02   \\ 
\cmidrule{1-5} \cmidrule{7-10} 
PIFS~\cite{cermelli2021prototype}                 & 28.78   & 3.82& 9.07   & 6.71 & \phantom{a}  & 17.84 & 2.87 & 6.02  & 4.88\\ 
attMPTI \cite{zhao2021few}     & 37.13 & 4.99 & 11.76 & 8.79 & \phantom{a}  & 54.84 & 3.28 & 14.14 & 6.17  \\ 
COSeg \cite{an2024rethinking} &57.67 & 5.21 & 16.25 & 9.54 & \phantom{a} & 47.03 & 4.03 & 13.09 & 7.42 \\
GW \cite{xu2023generalized} &59.28 & 8.30 & 19.03 & 14.55  & \phantom{a} & 55.23 & 6.47 & 16.74 & 11.56 \\
\gray\ourmodel~(ours) & \gray\textbf{67.57} & \gray\textbf{31.67} & \gray\textbf{39.23} & \gray\textbf{43.12}   & \gray\phantom{a} &\gray\textbf{68.48} &\gray\textbf{29.18} & \gray\textbf{37.45} & \gray\textbf{40.92} \\
\bottomrule
\end{tabular}}
\vspace{-2.53mm}
\caption{\textbf{Comparisons of our method with baselines on the new ScanNet200 benchmark.} The best results are highlighted in \textbf{bold}.}
\label{main-sc200}
\vspace{-1.57mm}
\end{table*}

\begin{table*}[t]
\centering
\renewcommand{\arraystretch}{.95}
\newcommand{\gray}{\cellcolor{mygray}}
\resizebox{0.98\linewidth}{!}{
\begin{tabular}{p{2.5cm}<\raggedright |p{1.5cm}<\centering p{1.5cm}<\centering p{1.5cm}<\centering p{1.5cm}<\centering p{0.01cm}<\centering p{1.5cm}<\centering p{1.5cm}<\centering p{1.5cm}<\centering p{1.5cm}<\centering}
\toprule
\multirow{2}{*}{Method} & \multicolumn{4}{c}{5-shot}  & \phantom{a} & \multicolumn{4}{c}{1-shot}   \\ \cmidrule{2-5} \cmidrule{7-10} 
                        & mIoU-B & mIoU-N & mIoU-A & HM & \phantom{a} & mIoU-B & mIoU-N & mIoU-A & HM \\ \midrule
Fully Supervised        & 65.45 & 37.24 & 48.53   & 47.47  & \phantom{a}   & 65.45 & 37.24 & 48.53   & 47.47   \\ 
\cmidrule{1-5} \cmidrule{7-10} 
PIFS \cite{cermelli2021prototype} & 39.98 & 5.74  & 19.44    & 10.03  & \phantom{a}   & 36.66      & 4.95       & 17.63     & 8.71 \\ 
attMPTI \cite{zhao2021few}          & 55.89      & 4.19       & 24.87      & 7.78   & \phantom{a}   & 53.16      & 3.55       & 23.40     & 6.66\\ 
COSeg~\cite{an2024rethinking}    & 59.34      & 6.96       &27.91      & 12.45     & \phantom{a}   & 58.49      & 6.24       & 27.14      & 11.26\\ 
GW \cite{xu2023generalized}   &51.35 & 11.03 & 27.16 & 18.15 & \phantom{a} & 46.71 & 6.63 & 22.66 & 11.59    \\
\gray\ourmodel~(ours)  & \gray\textbf{60.05} & \gray\textbf{21.66} & \gray\textbf{37.02} & \gray\textbf{31.82}   & \gray\phantom{a} &\gray\textbf{61.39} &\gray\textbf{19.42} & \gray\textbf{36.21} & \gray\textbf{29.47}              \\
\bottomrule
\end{tabular}}
\vspace{-2.53mm}
\caption{\textbf{Comparisons of our method with baselines on the new ScanNet++ benchmark.} The best results are highlighted in \textbf{bold}.}
\label{main-scpp}
\vspace{-1.57mm}
\end{table*}

\begin{table*}[!ht]
\centering
\renewcommand{\arraystretch}{.95}
\newcommand{\gray}{\cellcolor{mygray}}
\resizebox{0.98\linewidth}{!}{
\begin{tabular}{p{2.5cm}<\raggedright |p{1.5cm}<\centering p{1.5cm}<\centering p{1.5cm}<\centering p{1.5cm}<\centering p{0.01cm}<\centering p{1.5cm}<\centering p{1.5cm}<\centering p{1.5cm}<\centering p{1.5cm}<\centering}
\toprule
\multirow{2}{*}{Method} & \multicolumn{4}{c}{5-shot} & \phantom{a}  & \multicolumn{4}{c}{1-shot}   \\ \cmidrule{2-5} \cmidrule{7-10}  
                        & mIoU-B & mIoU-N & mIoU-A & HM & \phantom{a} & mIoU-B & mIoU-N & mIoU-A & HM \\ \midrule
Fully Supervised                    & 78.71      & 60.37       & 72.91      & 68.33 & \phantom{a}   & 78.71      & 60.37       & 72.91      & 68.33   \\ 
\cmidrule{1-5} \cmidrule{7-10} 
attMPTI \cite{zhao2021few}          & 16.31      & 3.12       & 12.35      & 5.21   & \phantom{a}  & 12.97      & 1.62       & 9.57     & 2.88\\ 
PIFS \cite{cermelli2021prototype}   & 35.14      & 3.21       &25.56      & 5.88    & \phantom{a}   & 35.80      & 2.54       & 25.82      & 4.75\\ 
CAPL \cite{tian2022generalized}   &38.22 & 14.39 & 31.07 & 20.88 & \phantom{a} & 38.70 & 10.59 & 30.27 & 16.53    \\
GW~\cite{xu2023generalized}   & 40.18 & 18.58 & 33.70 & 25.39 & \phantom{a}  &40.06 &14.78 & 32.47 & 21.55              \\
\gray\ourmodel~(ours)   & \gray\textbf{78.30} & \gray\textbf{51.22} & \gray\textbf{69.75} & \gray\textbf{61.91}  & \gray\phantom{a} & \gray\textbf{78.56} & \gray\textbf{49.72} & \gray\textbf{69.45} & \gray\textbf{60.88}              \\
\bottomrule
\end{tabular}
}
\vspace{-2.53mm}
\caption{\textbf{Comparisons of our method with baselines on the old ScanNet benchmark.} The best results are highlighted in \textbf{bold}.}
\label{main-scannet}
\vspace{-4.56mm}
\end{table*}

\begin{figure}[t!]
    \centering
    \includegraphics[width=.97\linewidth]{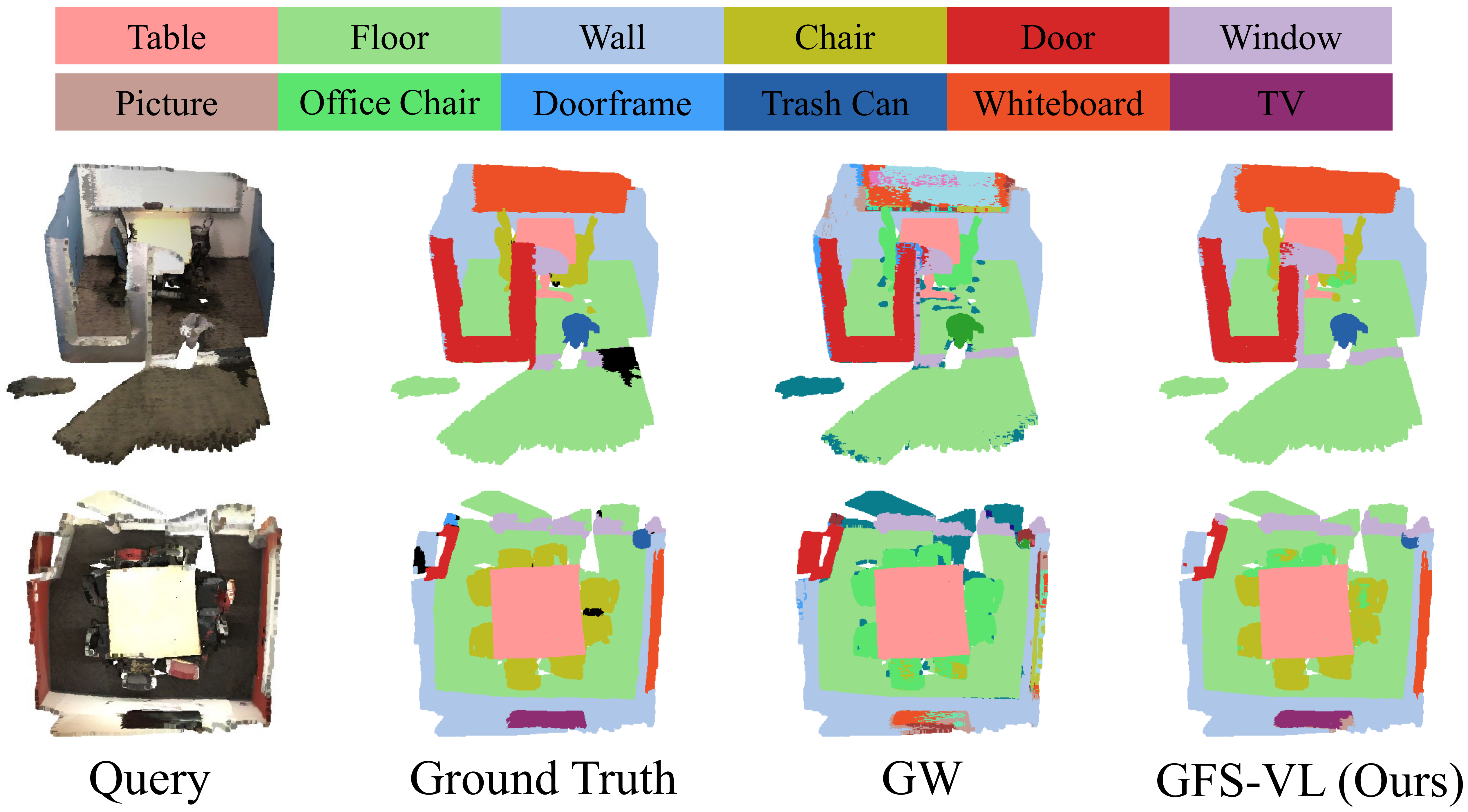}
    \vspace{-0.10in}
    \caption{
    \textbf{Qualitative comparison between GW~\cite{xu2023generalized} and our \ourmodel~on ScanNet200.} Class colors are shown at the top.
    }
    \label{fig:vs1}
    \vspace{-0.15in}
\end{figure}

\subsection{Experimental Results}
In our evaluation, we benchmark our method against baseline models attMPTI~\cite{zhao2021few} and PIFS~\cite{cermelli2021prototype} following~\cite{xu2023generalized}, along with the state-of-the-art GFS-PCS model GW~\cite{xu2023generalized} and FS-PCS model COSeg~\cite{an2024rethinking}. 
For fairness, all baseline models are retrained using the same backbone as our model. We also include a Fully Supervised model as an upper bound, obtained by fine-tuning an identical model to ours on ground-truth labels for both base and novel classes.

Following~\cite{tian2022generalized,xu2023generalized}, we evaluate two support scenarios, 5-shot and 1-shot, averaging performance across five randomly-seeded support set versions.
As shown in~\cref{main-sc200}, on ScanNet200, our model achieves substantial improvements across all metrics and support scenarios, 
including a 28.57\% increase in HM and a 23.37\% boost in mIoU-N over the closest baseline, GW, in the 5-shot setting. 
Qualitative comparisons in~\cref{fig:vs1} further illustrate the superior segmentation accuracy of our model compared to GW.
Similar trends are seen on ScanNet++ (see~\cref{main-scpp}), where our model improves HM by 17.88\% and mIoU-N by 12.79\% compared to GW (1-shot). 
For the traditional ScanNet benchmark with only six novel classes, we report results in~\cref{main-scannet} using baseline performance from~\cite{xu2023generalized}, where our model significantly surpasses all baselines with an impressive 34.94\% gain in mIoU-N and 39.33\% in HM for the 1-shot task.

The substantial and consistent gains across diverse datasets and metrics highlight our method's strong adaptability to diverse and complex novel classes by effectively integrating valuable few-shot samples with the semantic insights of 3D VLMs~\cite{ding2023pla,peng2023openscene}.
In contrast, baseline models rely solely on limited few-shot samples, constraining their capacity for novel class generalization and resulting in lower scores on the benchmarks. 
Moreover, the reduced performance on our new benchmarks emphasizes their value as a rigorous and comprehensive evaluation setting. These benchmarks challenge models to demonstrate robust generalization across diverse classes, fostering a deeper understanding of GFS-PCS for real-world applications.

\begin{table*}[!t]
\begin{subtable}{0.35\linewidth}
    \centering
    \renewcommand{\arraystretch}{0.833}
    \setlength{\tabcolsep}{3pt}
    \resizebox{\linewidth}{!}{
    \begin{tabular}{@{}lccccccc@{}} 
    \toprule
     PS   &  AI  &  NB-Mix & mIoU-B & mIoU-N & mIoU-A & HM\\ \midrule 
                 &            &            &    65.50  & 22.30 & 31.40  & 33.28  \\
    \checkmark   &            &            &    69.26    & 26.51 & 35.51 & 38.35  \\
    \checkmark   & \checkmark &            &    66.25    & 28.03 & 36.07 & 39.39  \\
    \checkmark   &            & \checkmark &    66.94    & 28.21 & 36.36 & 39.69  \\
    \checkmark   & \checkmark & \checkmark &    67.42    & 31.81 & 39.30 & 43.22  \\
    \bottomrule
    \end{tabular}}
    \subcaption{}
    \label{table:design}
\end{subtable}
\hfill
\begin{subtable}{.35\linewidth}
    \centering
    \renewcommand{\arraystretch}{1.048}
    \setlength{\tabcolsep}{3pt}
    \resizebox{\linewidth}{!}{
    \begin{tabular}{@{}lccccc@{}} 
    \toprule
     3D VLM     & mIoU-B & mIoU-N & mIoU-A & HM \\ 
                                      \midrule 
        RegionPLC~\cite{yang2024regionplc}   & 46.97  & 23.77 & 28.65 & 31.56 \\ 
        Ours (RegionPLC)                    &  67.42 & 31.81 & 39.30 & 43.22 \\ 
        Openscene~\cite{peng2023openscene}    &  53.07 & 15.16 & 23.14 & 23.58 \\ 
        Ours (Openscene)                       &  68.56 & 20.09 & 30.29 & 31.07 \\ 
    \bottomrule
    \end{tabular}}
    \subcaption{}
    \label{table:vlms}
\end{subtable}
\hfill
\begin{subtable}{0.27\linewidth}
    \centering
    \renewcommand{\arraystretch}{1}
    \setlength{\tabcolsep}{3pt}
    \resizebox{\linewidth}{!}{
    \begin{tabular}{@{}lccccc@{}} 
    \toprule
     $\delta$     & mIoU-B & mIoU-N & mIoU-A & HM \\ 
                                      \midrule 
       0.80    &  66.67 & 30.41 & 38.04 & 41.77 \\ 
       0.85     &  66.29 & 31.00 & 38.43 & 42.24 \\ 
       0.90     &  67.42 & 31.81 & 39.30 & 43.22 \\ 
       0.95     & 66.22  & 30.68 & 38.17 & 41.94 \\ 
    \bottomrule
    \end{tabular}}
    \subcaption{}
    \label{table:flth}
\end{subtable}
\hfill
\begin{subtable}{0.232\linewidth}
    \vspace{0.05in}
    \centering
    \renewcommand{\arraystretch}{.9}
    \setlength{\tabcolsep}{2.4pt}
    \resizebox{\linewidth}{!}{
    \begin{tabular}{@{}lccccc@{}} 
    \toprule
     $\tau$     & mIoU-B & mIoU-N & mIoU-A & HM \\ 
                                      \midrule 
      0.5      & 67.19  & 31.33 & 38.88 & 42.73 \\ 
      0.6      & 67.42 & 31.81 & 39.30 & 43.22 \\ 
      0.7      & 66.69 & 30.56 & 38.17 & 41.92  \\ 
      0.8      &  68.06 & 30.67 & 38.54 & 42.28 \\ 
    \bottomrule
    \end{tabular}}
    \subcaption{}
    \label{table:psth}
\end{subtable}
\hfill
\begin{subtable}{0.222\linewidth}
    \centering
    \renewcommand{\arraystretch}{.9}
    \setlength{\tabcolsep}{2.4pt}
    \resizebox{\linewidth}{!}{
    \begin{tabular}{@{}lccccc@{}} 
    \toprule
     $n$     & mIoU-B & mIoU-N & mIoU-A & HM \\ 
                                      \midrule 
       1     & 67.40  & 27.13 & 35.61 & 38.68 \\
       2     &  67.95 & 27.71 & 36.18 & 39.36 \\ 
       3     &  66.94 & 28.21 & 36.36 & 39.69 \\ 
       4     &  67.84 & 27.80 & 36.23 & 39.44 \\ 
    \bottomrule
    \end{tabular}}
    \subcaption{}
    \label{table:mixup}
\end{subtable}
\hfill
\begin{subtable}{.26\linewidth}
    \centering
    \renewcommand{\arraystretch}{1.45}
    \setlength{\tabcolsep}{2.4pt}
    \resizebox{\linewidth}{!}{
    \begin{tabular}{@{}lccccc@{}} 
    \toprule
     Backbone     & mIoU-B & mIoU-N & mIoU-A & HM \\ 
                                      \midrule 
        PTv3~\cite{wu2024point}    &  67.42 & 31.81 & 39.30 & 43.22 \\ 
        SCN~\cite{graham20183d}    &  61.85 & 31.94 & 38.24 & 42.13 \\ 
    \bottomrule
    \end{tabular}}
    \subcaption{}
    \label{table:backbone}
\end{subtable} 
\hfill
\begin{subtable}{.268\linewidth} 
    \centering
    \renewcommand{\arraystretch}{1.13}
    \setlength{\tabcolsep}{2.4pt}
    \resizebox{\linewidth}{!}{
    \begin{tabular}{@{}lccccc@{}} 
    \toprule
     Mix     & mIoU-B & mIoU-N & mIoU-A & HM \\  \midrule 
       Instance Mix    & 68.29  & 23.93 & 33.27 & 35.44 \\ 
       Mix3D~\cite{nekrasov2021mix3d} & 68.50  & 24.80 &34.00 & 36.42 \\ 
       NB-Mix   & 67.95  & 27.71 & 36.18 & 39.36 \\ 
    \bottomrule
    \end{tabular}}
    \subcaption{}
    \label{table:mixdesign}
\end{subtable} 
\vspace{-0.25in}
\caption{\textbf{Ablation study.} (a) Effect of design components. (b) Results with different 3D VLMs. (c) Impact of the threshold in AI. (d) Effect of the threshold in PS. (e) Impact of different mix blocks. (f) Results with different backbones. (g) Comparison of mix strategies.}
\vspace{-0.07in}
\end{table*}

\begin{figure*}[t!]
    \centering
    \includegraphics[width=.95\linewidth]{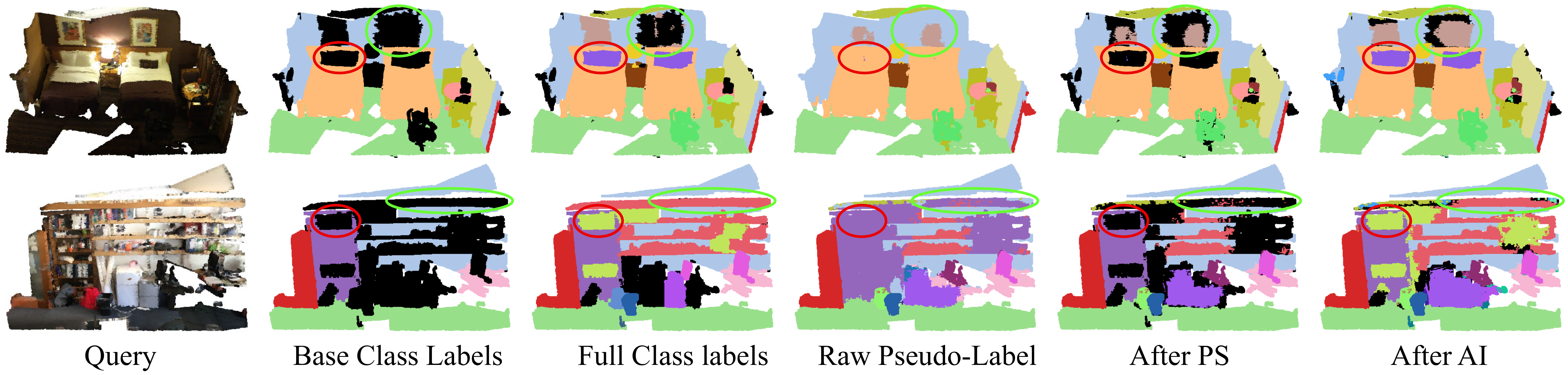}
    \vspace{-0.10in}
    \caption{
    Visualization of the improvements in pseudo-label quality after applying Pseudo-label Selection (PS) and Adaptive Infilling (AI). Note that AI effectively discovers missed novel classes in the red circles and completes partial pseudo-labels in the green circles.
    }
    \label{fig:vs2}
    \vspace{-0.15in}
\end{figure*}

\begin{figure}[t!]
    \centering
    \includegraphics[width=.95\linewidth]{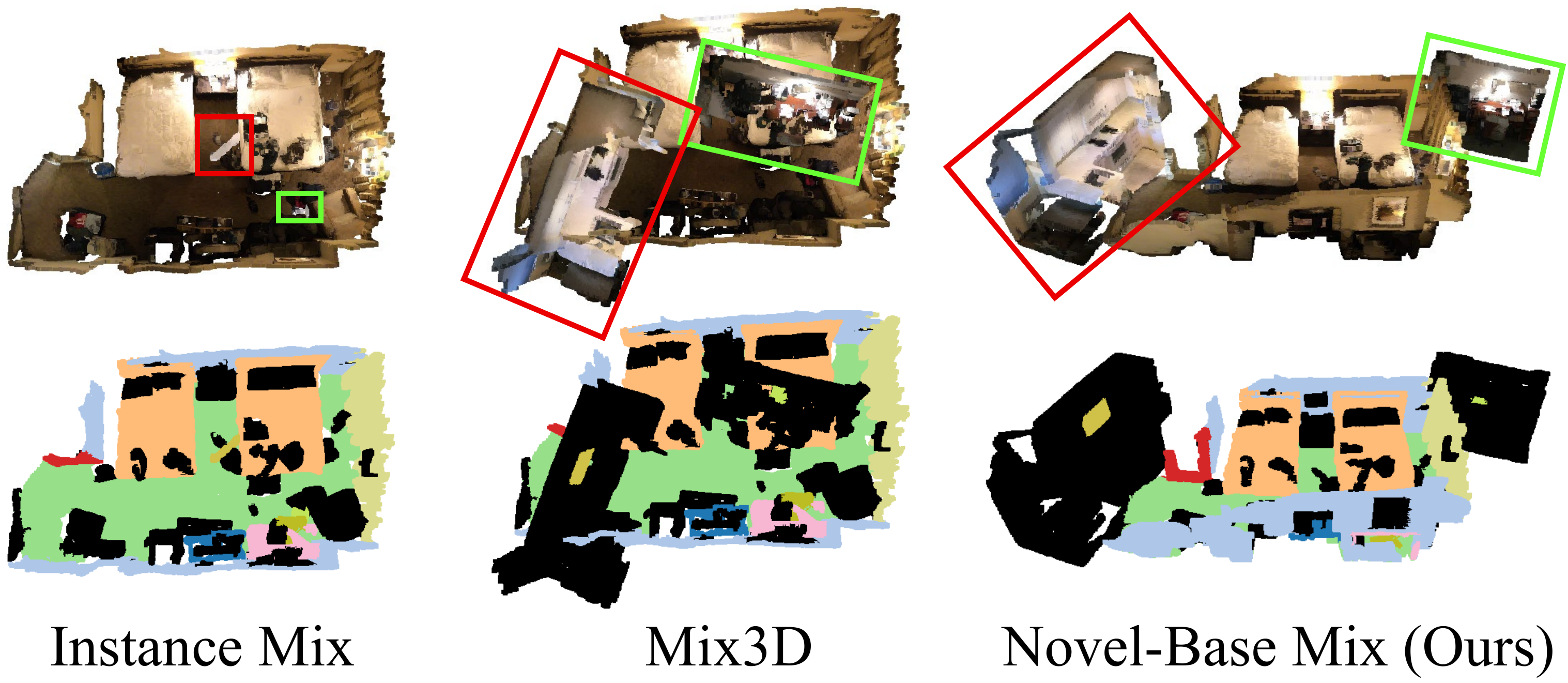}
    \vspace{-0.10in}
    \caption{
    Visual illustration of mixing strategies. 
    The red and green boxes represent the two novel samples mixed into the scene.
    }
    \label{fig:vs3}
    \vspace{-0.15in}
\end{figure}

\subsection{Ablation Studies}
\label{sec:abl}
We conducted ablation studies on the ScanNet200~\cite{rozenberszki2022language} dataset using a single set of 5-shot support samples.

\noindent\textbf{Effect of Design Components.} 
We assessed the effectiveness of each module in~\ourmodel~in~\cref{table:design}. 
The baseline (1$^\text{st}$ row) shows performance using raw pseudo-labels directly from the 3D VLM. 
Adding Pseudo-label Selection (PS) in the 2$^\text{nd}$ row significantly improves pseudo-label quality, resulting in a clear performance boost.
Introducing Adaptive Infilling (AI) further enhances results by effectively assigning labels to unlabeled regions of novel classes (3$^\text{th}$ row). Visualizations in~\cref{fig:vs2} illustrate the quality improvements achieved with PS and AI, affirming their effectiveness.
Lastly, Novel-Base Mix (NB-Mix), whether used with PS (4$^\text{rd}$ row) or combined with PS and AI (5$^\text{th}$ row), improves generalization by effectively integrating novel knowledge into the training samples.

\noindent\textbf{Results with Different 3D VLMs.}
We tested~\ourmodel with two prominent 3D VLMs, RegionPLC~\cite{yang2024regionplc} and Openscene~\cite{peng2023openscene}, as shown in the 2$^\text{nd}$ and 4$^\text{th}$ rows of~\cref{table:vlms}. 
Compared to their zero-shot results (1$^\text{st}$ and 3$^\text{rd}$ rows), our approach achieves substantial gains by effectively leveraging few-shot samples to refine and enhance the noisy knowledge in each 3D VLM. This demonstrates our framework’s flexibility and effectiveness across diverse 3D VLMs.

\noindent\textbf{Impact of the Threshold in AI.} 
Adaptive Infilling (AI) assigns novel class labels to unlabeled points based on similarity to adaptive prototypes, controlled by a threshold $\delta$. 
In~\cref{table:flth}, we explore different $\delta$ values and observe that $\delta = 0.9$ achieves the highest performance, which best balances enriching novel classes and maintaining label quality.

\noindent\textbf{Effect of the Threshold in PS.} 
Pseudo-label Selection (PS) refines pseudo-labels by retaining only the most reliable regions predicted by the 3D VLM, based on a threshold $\tau$. 
In~\cref{table:psth}, we evaluate the effect of varying $\tau$. 
The method performs robustly across different $\tau$ values, with the optimal performance achieved at $\tau = 0.6$. This robustness indicates that PS effectively selects relevant reliable regions without being overly sensitive to the threshold choice.

\noindent\textbf{Impact of Different Mix Blocks.} 
In NB-Mix, we can adjust the number of novel blocks, denoted as $n$, used to augment each training sample. 
\cref{table:mixup} shows stable performance (with the AI module disabled) across different block counts, suggesting that NB-Mix effectively integrates novel class information.
By default, we use three blocks.

\noindent\textbf{Results with Different Backbones.} 
Besides varying 3D VLMs, we examined the effect of different backbones. 
We evaluated two backbones, PTv3~\cite{wu2024point} and SparseConvNet (SCN)~\cite{graham20183d} in~\cref{table:backbone}. 
Our approach consistently performs well across both networks, 
confirming that~\ourmodel~is generalizable and not dependent on a specific backbone.

\noindent\textbf{Comparison of Mix Strategies.} 
We investigated alternative mix strategies for integrating novel support samples into training data in~\cref{table:mixdesign}. 
Specifically, we compared our NB-Mix with Instance Mix, which randomly inserts novel class objects from foreground masks into scenes, and Mix3D~\cite{nekrasov2021mix3d}, which overlays two scenes for out-of-context augmentation. 
\cref{fig:vs3} shows the visual examples of these strategies.
Our method outperforms these alternatives, highlighting the importance of preserving local context to effectively learn diverse and challenging novel classes.

\section{Conclusion}
This work introduces a GFS-PCS framework~\ourmodel~that synergizes dense but noisy pseudo-labels from 3D VLMs with accurate yet sparse few-shot samples, overcoming current GFS-PCS limitations in novel knowledge learning.
\ourmodel~utilizes prototype-guided pseudo-label selection to target high-quality regions and adaptive infilling to enrich pseudo-labels. 
Besides, the novel-base mix embeds few-shot samples into training scenes, preserving essential context for improved novel class learning. 
Identifying the limited diversity in current GFS-PCS evaluations, we introduce two benchmarks with broader, more diverse novel classes for more comprehensive generalization evaluation. 
\ourmodel~achieves leading results and generalizes effectively across models and datasets, showing the potential of 3D VLMs in advancing GFS-PCS. 
We hope our method and benchmarks serve as a foundation for future research.

\section*{Acknowledgments} This work is supported by the Pioneer Centre for AI, DNRF grant number P1.

{
    \small
    \bibliographystyle{ieeenat_fullname}
    \bibliography{main}
}

\clearpage
\setcounter{page}{1}
\maketitlesupplementary

\begin{algorithm}[!t]
\caption{\small{Pseudo-code of Novel-Base Mix in PyTorch style.}}
\definecolor{codeblue}{rgb}{0.25,0.5,0.5}
\definecolor{codegreen}{rgb}{0,0.6,0}
\definecolor{codegray}{rgb}{0.5,0.5,0.5}
\definecolor{codepurple}{rgb}{0.58,0,0.82}
\definecolor{backcolour}{rgb}{0.95,0.95,0.92}
\lstset{
  backgroundcolor=\color{white},
  basicstyle=\fontsize{7.2pt}{7.2pt}\ttfamily\selectfont,
  columns=fullflexible,
  breaklines=true,
  captionpos=b,
  commentstyle=\fontsize{7.2pt}{7.2pt}\color{codeblue},
  numberstyle=\tiny\color{codegray},
  stringstyle=\color{codepurple},
  keywordstyle=\fontsize{7.2pt}{7.2pt}\color{magenta},
}
\begin{lstlisting}[language=python]
# Input:
# novel_cloud: point cloud of the novel class, shape (N, 3)
# novel_mask: binary mask for points belonging to the novel class, shape (N,)
# base_cloud: point cloud of base classes, shape (M, 3)
# random_corner: function to randomly select a corner ('bottom', 'top', 'left', 'right')
# crop_fn: function to crop novel point clouds based on the mask
# corner_fn: function to compute corner points in the XY plane, 
#            returning a dictionary with keys: ['top', 'bottom', 'left', 'right']

# Step 1: Crop the novel point cloud based on the mask
novel_local, novel_local_mask = crop_fn(novel_cloud, novel_mask)

# Step 2: Compute corner points for both point clouds
base_corners = corner_fn(base_cloud) 
novel_corners = corner_fn(novel_local)

# Step 3: Randomly select a corner for alignment
selected_corner = random_corner(['bottom', 'top', 'left', 'right'])

# Step 4: Calculate the translation vector based on selected corners
if selected_corner == "bottom":
    base_point = base_corners['bottom']      # Lowest point of base cloud in Y
    novel_point = novel_corners['top']       # Highest point of novel cloud in Y
elif selected_corner == "top":
    base_point = base_corners['top']         # Highest point of base cloud in Y
    novel_point = novel_corners['bottom']    # Lowest point of novel cloud in Y
elif selected_corner == "left":
    base_point = base_corners['left']        # Leftmost point of base cloud in X
    novel_point = novel_corners['right']     # Rightmost point of novel cloud in X
else:  # "right"
    base_point = base_corners['right']       # Rightmost point of base cloud in X
    novel_point = novel_corners['left']      # Leftmost point of novel cloud in X

translation_vector = [base_point[0] - novel_point[0], 
                      base_point[1] - novel_point[1], 
                      0]  # No z-translation yet

# Step 5: Translate the novel cloud in the XY plane
novel_local_translated = novel_local + translation_vector

# Step 6: Align the z-coordinates
z_adjustment = min(base_cloud[:, 2]) - min(novel_local_translated[:, 2])
novel_local_translated[:, 2] += z_adjustment

# Step 7: Combine base cloud and translated novel cloud
mixed_cloud = torch.cat([base_cloud, novel_local_translated], dim=0)
\end{lstlisting}
\label{alg:nbmix}
\end{algorithm}

\section{Additional Details on Novel-Base Mix}

To effectively utilize support samples, the Novel-Base Mix approach is designed to integrate them into the base training inputs while preserving essential scene context. 
This ensures effective learning of challenging novel classes. 
We provide the pseudo-code for Novel-Base Mix in~\cref{alg:nbmix}. Below, we present a step-by-step explanation of the process.

\myPara{Step 1.}
The process begins with cropping the region of novel objects from the novel point cloud. Given the \textit{randomly sampled} novel point cloud and its corresponding binary mask, a cropping operation is applied to extract the relevant local region. 
This ensures that only the novel object region is considered for mixing, while extraneous unnecessary points are excluded.

\myPara{Step 2.}
Next, to align the cropped novel sample with the base point cloud, we identify key spatial anchors in the \textit{XY plane}.
These anchors correspond to the top, bottom, left, and rightmost corner points of both the base point cloud and the cropped novel point cloud. These anchors serve as reference points for spatial alignment in subsequent steps.

\begin{figure*}[t!]
    \centering
    \includegraphics[width=\linewidth]{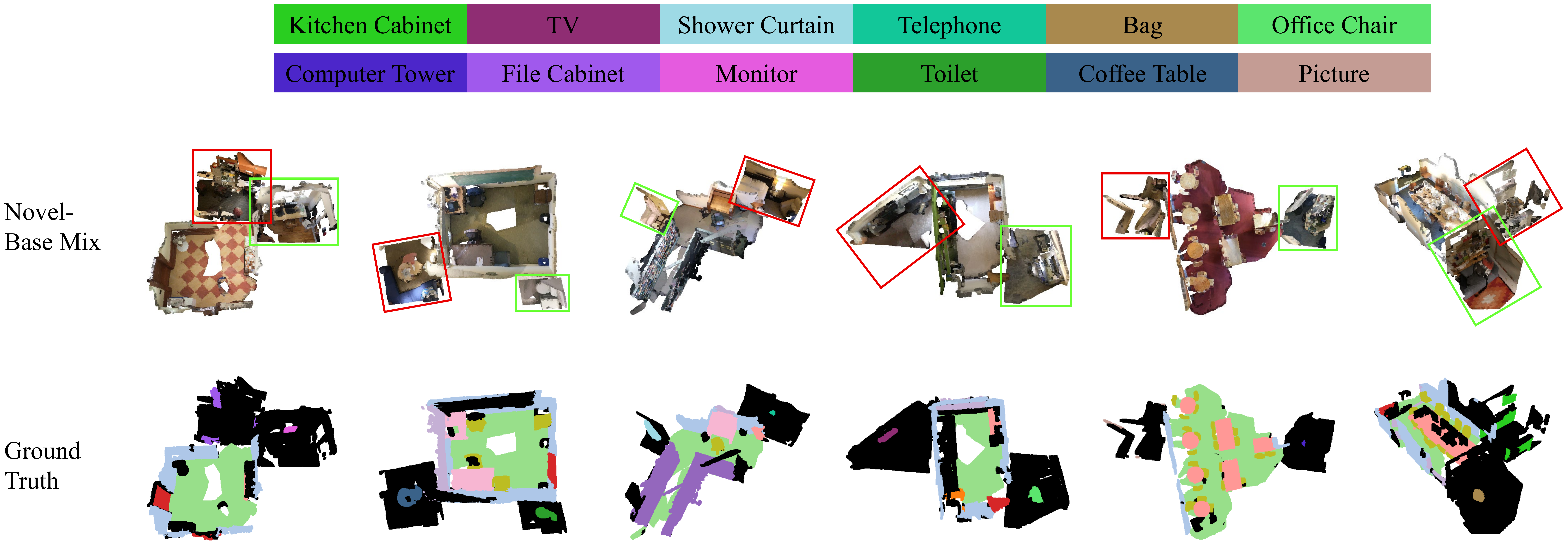}
    \caption{
    Visualization of the outputs from the proposed Novel-Base Mix. The red and green boxes represent the two novel samples mixed into the scene. The novel class colors are shown at the top. 
    }
    \label{fig:morenb}
\end{figure*}

\myPara{Step 3–4.}
A random corner from \texttt{\{top, bottom, left, right\}} is selected for alignment.
For instance:
\begin{itemize}
\item If the bottom corner is chosen, the lowest corner of the base point cloud is aligned with the highest corner of the novel point cloud.
\item Conversely, if the left corner is selected, the leftmost corner of the base point cloud is aligned with the rightmost corner of the novel point cloud.
\end{itemize}

This pairing strategy introduces diversity in the placement of novel objects while ensuring the preservation of contextual integrity.
Based on the selected corner pair, a translation vector is computed to spatially align the novel sample with the base point cloud.

\myPara{Step 5–6.}
The computed translation vector is applied to the cropped novel point cloud, ensuring that it is positioned next to the base point cloud in the XY plane.

Additionally, Z-axis alignment is performed by adjusting the Z-coordinates of the translated novel point cloud. This step ensures that the novel sample is grounded at the same level as the base point cloud, preventing it from floating above or sinking below the base scene.

\myPara{Step 7.}
Finally, the aligned novel sample is merged with the base point cloud to form the mixed training input. This effectively integrates the novel sample into the training scene while retaining its original context, which helps the model recognize complex and challenging novel classes.

We provide additional visualizations of the outputs from our Novel-Base Mix in~\cref{fig:morenb}.

\section{Additional Details on the new Benchmarks}
\label{sec:id}
As discussed in Sec.\textcolor{cvprblue}{5.1}, we leverage two recent datasets, ScanNet200~\cite{rozenberszki2022language} and ScanNet++~\cite{yeshwanth2023scannet++}, to construct comprehensive evaluation benchmarks for GFS-PCS.

ScanNet200~\cite{rozenberszki2022language} extends the labeling space of ScanNet~\cite{dai2017scannet} from 20 to 200 categories, introducing finer-grained subclasses of existing categories and numerous novel object types. 
These expansions enhance the dataset's granularity and diversity, making it a valuable resource for evaluating GFS-PCS methods. 
Meanwhile, ScanNet++~\cite{yeshwanth2023scannet++} offers annotations for 460 scenes encompassing over 1,000 unique object classes.
This dataset captures a broad range of object categories, reflecting the complexity and variability of real-world environments. 
Together, these datasets form a rich and diverse foundation for constructing robust GFS-PCS evaluation benchmarks.

\myPara{Benchmark Design.}
To create meaningful and representative GFS-PCS benchmarks, we carefully selected classes based on their occurrence counts in the respective datasets, ensuring sufficient representation across scenes. 
The selection process involved computing the occurrence count of each class across the dataset, ranking the classes by their occurrence counts, and assigning them to base and novel sets as follows:

\begin{itemize}
    \item ScanNet200: 
    Classes with occurrence counts exceeding 100 were retained, yielding 57 classes in total. The 12 most frequently occurring classes were designated as base classes, while the remaining 45 were assigned as novel classes.
    \item ScanNet++:
    Classes with occurrence counts exceeding 80 were retained, resulting in 30 classes in total. The top 12 most frequent classes formed the base class set, while the remaining 18 were assigned as novel classes.
\end{itemize}

These frequency thresholds were carefully chosen to strike a balance between class diversity and adequate representation, ensuring that both base and novel classes are well-suited for evaluating GFS-PCS performance.

\myPara{Benchmark Classes.}
The following are the specific class lists for the two benchmarks:

\begin{itemize}
    \item ScanNet200:
    \begin{itemize}
    \item \texttt{Base Classes}: [`refrigerator', `desk', `curtain', `bookshelf', `bed', `table', `window', `cabinet', `door', `chair', `floor', `wall']
    \item \texttt{Novel Classes}: [`trash can', `ceiling', `doorframe', `object', `shelf', `sink', `picture', `backpack', `couch', `box', `pillow', `radiator', `mirror', `whiteboard', `lamp', `toilet', `book', `monitor', `towel', `tv', `clothes', `coffee table', `office chair', `nightstand', `bag', `dresser', `toilet paper', `recycling bin', `kitchen cabinet', `bathtub', `telephone', `plant', `stool', `keyboard', `shoe', `jacket', `shower curtain', `armchair', `microwave', `computer tower', `bathroom vanity', `kitchen counter', `shower wall', `paper towel dispenser', `file cabinet']
    \end{itemize}
\end{itemize}

\begin{itemize}
    \item ScanNet++:
    \begin{itemize}
    \item \texttt{Base Classes}: [`wall', `floor', `door', `ceiling', `table', `window', `box', `ceiling lamp', `light switch', `cabinet', `chair', `heater']
    \item \texttt{Novel Classes}: [`monitor', `whiteboard', `office chair', `bottle', `doorframe', `keyboard', `window frame', `mouse', `paper', `blinds', `trash can', `telephone', `book', `shelf', `sink', `windowsill', `bag', `smoke detector']
    \end{itemize}
\end{itemize}

Overall, our benchmarks provide a more robust and comprehensive testbed for evaluating GFS-PCS methods. 
By better reflecting real-world challenges, our benchmarks enable researchers to rigorously assess models' performance and generalization to novel categories under realistic scenarios.

\section{Additional Implementation Details}
Our framework employs a straightforward segmentor consisting of a backbone and a linear classification head, designed for both efficiency and simplicity to facilitate reproducibility. 
The training process consists of two stages: pretraining on the base classes of each dataset, followed by fine-tuning with adding a separate linear classification head for novel classes.
For prompting the 3D VLMs, we adopt the default prompt used in RegionPLC~\cite{yang2024regionplc} and OpenScene~\cite{peng2023openscene}: ``a CLASS\_NAME in a scene".
We evaluate two widely used backbones in our experiments: Point Transformer V3 (PTv3)~\cite{wu2024point} and SparseConvNet (SCN)~\cite{graham20183d}. 
All experiments were conducted using 4 NVIDIA RTX 4090 GPUs.

For pretraining, we adhere to the default configurations provided in~\cite{wu2024point}. When using PTv3 as the backbone, the model is trained for 800 epochs with the AdamW optimizer. The learning rate is set to 0.006, with a reduced learning rate of 0.0006 for the backbone blocks, and a weight decay of 0.05. The OneCycleLR scheduler is employed to adjust the learning rate during training.
When using SCN as the backbone, the model is trained for 600 epochs on ScanNet200 and 800 epochs on ScanNet++ and ScanNet. The SGD optimizer is employed, with a learning rate of 0.05 and a weight decay of 0.0001. Similar to PTv3, the learning rate is scheduled using the OneCycleLR strategy.

For fine-tuning, the network is trained for 20 epochs end-to-end with the Adam optimizer. A constant learning rate is used, with a value of 0.001 for ScanNet200 and ScanNet, and 0.007 for ScanNet++. The backbone learning rate is reduced by a factor of 0.1 to stabilize training.

During training, the preprocessing follows the steps outlined in~\cite{wu2024point}. The raw input points are voxelized using a grid size of 0.02m, and a random cropping operation is applied to ensure that the number of points in each training input remains within a maximum limit, such as 102,400 points.

In the evaluation phase, the input point clouds only undergo voxelization without any further cropping or sampling operations. 
This enables testing on full scenes, as opposed to small blocks used in previous GFS-PCS evaluations~\cite{xu2023generalized, zhao2021few}.
By evaluating on entire scenes, it better simulates real-world scenarios and provides a more realistic and comprehensive assessment of models' performance.

\section{Additional Visualizations}
In this section, we present additional qualitative results to further illustrate the efficacy of our approach in addressing GFS-PCS tasks. These results highlight the superiority of our model in novel class generalization and segmentation quality, providing deeper insights into the design and impact of our proposed modules.

\begin{figure*}[t!]
    \centering
    \includegraphics[width=\linewidth]{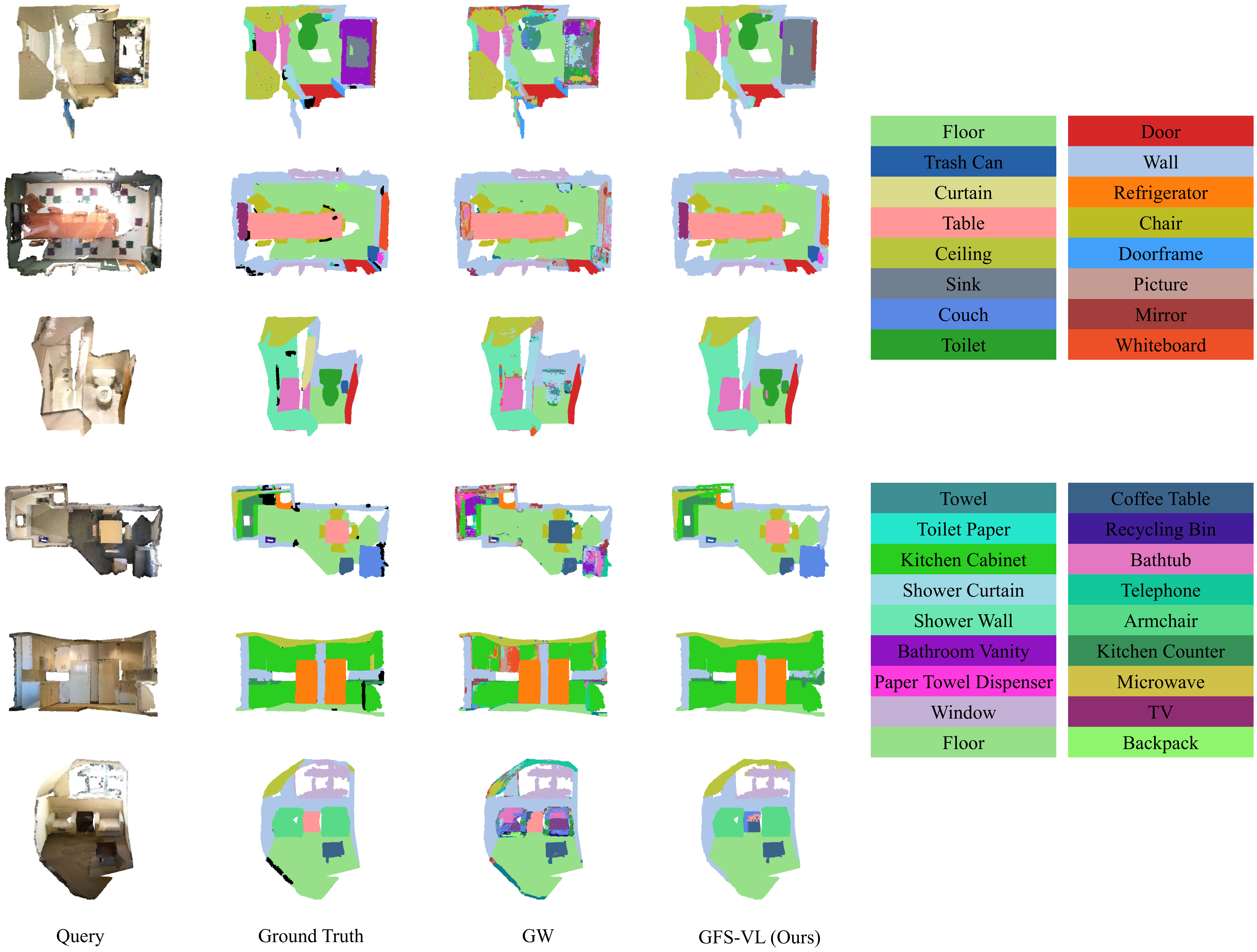}
    \caption{
    \textbf{Qualitative comparison between GW~\cite{xu2023generalized} and our \ourmodel~on ScanNet200.} 
    The visualizations demonstrate the superior segmentation performance and novel class generalization capabilities of~\ourmodel. 
    For clarity, class colors are displayed on the right and are restricted to those present in the ground truth annotations.
    }
    \label{fig:morepre}
\end{figure*}

\myPara{Comparison with State-of-the-Art Methods.}
Figure~\ref{fig:morepre} showcases additional segmentation results comparing our proposed framework,~\ourmodel, against the previously established state-of-the-art method, GW~\cite{xu2023generalized}, on the ScanNet200~\cite{rozenberszki2022language} benchmark. 
For clarity, class colors used in the visualizations are displayed on the right side of the figure and are restricted to those present in the ground truth.

These visualizations clearly demonstrate the superior performance of~\ourmodel, which effectively integrates dense semantic knowledge embedded in 3D VLMs with precise guidance from few-shot samples. 
This synergy enables~\ourmodel~to achieve robust novel class generalization in the challenging benchmarks. The qualitative results highlight improved boundary delineation, more accurate segmentation of novel objects, and a better overall alignment with ground truth.

Despite achieving better performance, Figure~\ref{fig:morepre} also exposes some limitations. Specifically, our model exhibits suboptimal performance on small objects (\eg, \textit{Trash Can} in the third row), thin objects (\eg, \textit{Curtain} in the third row), and objects within complex backgrounds (\eg, \textit{Bathroom Vanity} in the first row). Addressing these challenges presents promising directions for future work.

\begin{figure*}[t!]
    \centering
    \includegraphics[width=\linewidth]{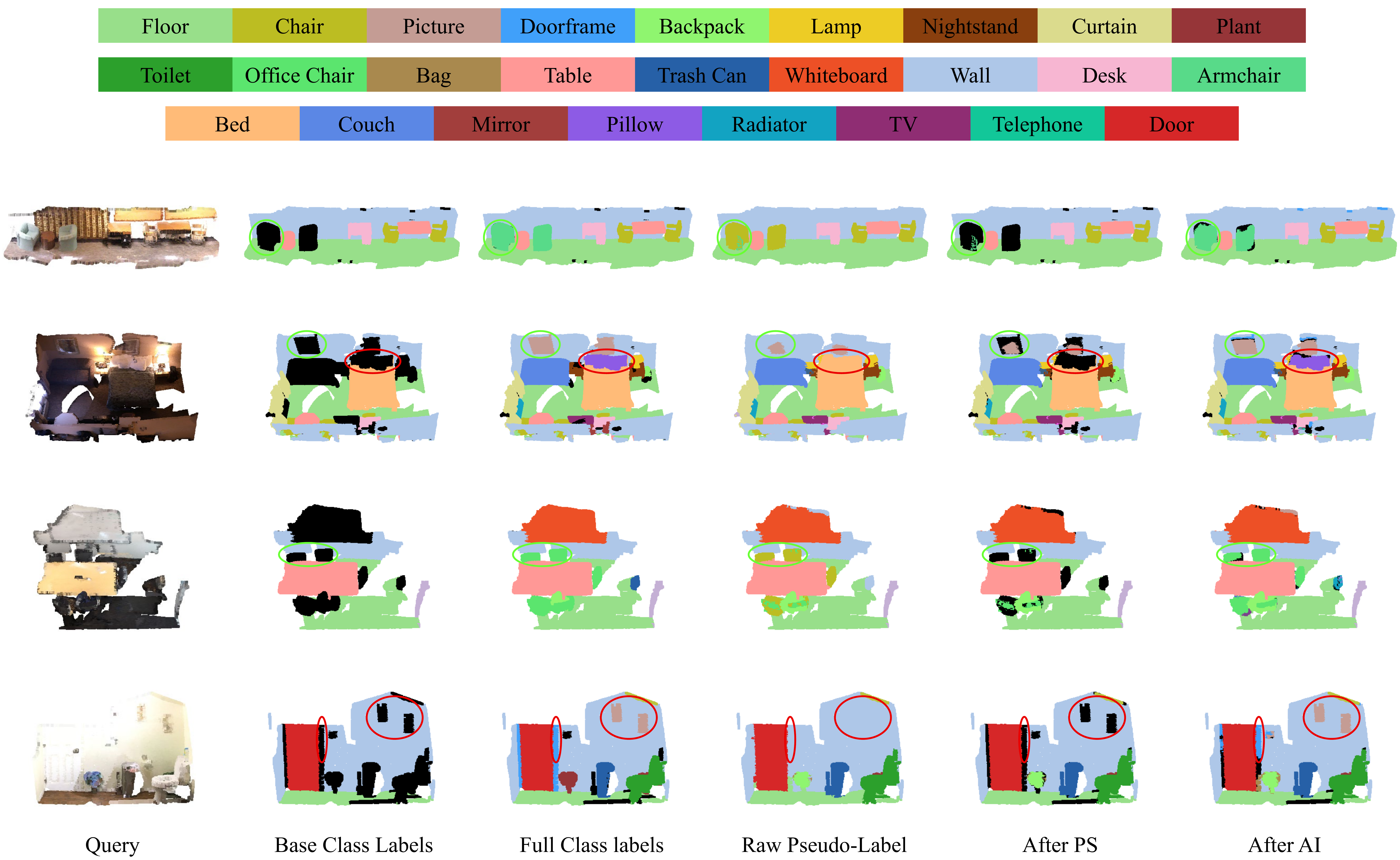}
    \caption{
    \textbf{Visualization of pseudo-label refinement using Pseudo-label Selection (PS) and Adaptive Infilling (AI).}  
    Red circles indicate novel objects discovered by AI that were missed in the raw pseudo-labels, while green circles indicate regions where AI completes previously partially segmented areas. 
    For clarity, class colors are displayed at the top and correspond to labels present in the full class annotations.
    }
    \label{fig:moreps}
\end{figure*}

\myPara{Improvements in Pseudo-label Quality.}
In Figure~\ref{fig:moreps}, we provide additional visualizations of the refinement process for raw pseudo-labels, illustrating the role of our Pseudo-label Selection (PS) and Adaptive Infilling (AI) modules. 
\begin{itemize}
    \item PS filters noisy predictions from the 3D VLM by anchoring them to the accurate few-shot samples, ensuring high reliability.
    \item AI discovers novel objects that were initially missed in the raw pseudo-labels, as indicated in the red circles in Figure~\ref{fig:moreps}, and completes partially segmented regions, as shown in the green circles. 
\end{itemize}

By integrating few-shot support samples with the current pseudo-label context, the AI module creates adaptive prototypes that facilitate both the discovery of missed novel objects and the completion of partial pseudo-labels,  thereby enhancing the quality of novel region labels. Together, PS and AI play distinct yet complementary roles in pseudo-label refinement. By combining dense knowledge from 3D VLMs with the precision of few-shot samples, these modules significantly improve pseudo-label quality, achieving better alignment with the full-class ground truth.

For visualization, the class colors in Figure~\ref{fig:moreps} are displayed at the top and correspond to labels present in the full-class annotations.

\end{document}